\documentclass{article}

\usepackage{arxiv}
\usepackage{comment}

\usepackage{algorithm}
\usepackage{algpseudocode}

\usepackage[utf8]{inputenc} % allow utf-8 input
\usepackage[T1]{fontenc}    % use 8-bit T1 fonts
\usepackage{hyperref}       % hyperlinks
\usepackage{url}            % simple URL typesetting
\usepackage{booktabs}       % professional-quality tables
\usepackage{amsfonts}       % blackboard math symbols
\usepackage{nicefrac}       % compact symbols for 1/2, etc.
\usepackage{microtype}      % microtypography
\usepackage{lipsum}
\usepackage{graphicx}
\newcommand{\beq}{\begin{eqnarray}}
\newcommand{\eeq}{\end{eqnarray}}

\title{Iterative CT Reconstruction via Latent Variable Optimization of Shallow Diffusion Models}

\author{
 Sho Ozaki \\
  Graduate School of Science and Technology\\
  Hirosaki University\\
  3 Bunkyo, Hirosaki, Aomori 036-8561, Japan \\
  \texttt{sho.ozaki@hirosaki-u.ac.jp} \\
   \And
 Shizuo Kaji \\
  Institute of Mathematics for Industry\\
  Kyushu University\\
  744 Motooka, Nishi-ku, Fukuoka 819-0395, Japan \\
  \AND
  Toshikazu Imae \\
  Department of Radiology\\
  University of Tokyo Hospital\\
  Tokyo 113-8655, Japan \\
  \AND
  Kanabu Nawa\\
  Kansai BNCT Medical Center \\
  Osaka Medical and Pharmaceutical University \\ 
  2-7 Daigaku-machi, Takatsuki-shi \\
  Osaka 569-8686, Japan \\
  \And
  Hideomi Yamashita \\
  Department of Radiology\\
  University of Tokyo Hospital\\
  Tokyo 113-8655, Japan \\
  \And
  Keiichi Nakagawa \\
  Department of Radiology\\
  University of Tokyo Hospital\\
  Tokyo 113-8655, Japan \\
}
\date{}

\begin{document}
\fontsize{11}{14}\selectfont
\maketitle
\begin{abstract}
Image-generative artificial intelligence (AI) has garnered significant attention in recent years. In particular, the diffusion model, a core component of generative AI, produces high-quality images with rich diversity. In this study, we proposed a novel computed tomography (CT) reconstruction method by combining the denoising diffusion probabilistic model with iterative CT reconstruction. In sharp contrast to previous studies, we optimized the fidelity loss of CT reconstruction with respect to the latent variable of the diffusion model, instead of the image and model parameters. To suppress the changes in anatomical structures produced by the diffusion model, we shallowed the diffusion and reverse processes and fixed a set of added noises in the reverse process to make it deterministic during the inference. We demonstrated the effectiveness of the proposed method through the sparse-projection CT reconstruction of 1/10 projection data. Despite the simplicity of the implementation, the proposed method has the potential to reconstruct high-quality images while preserving the patient's anatomical structures and was found to outperform existing methods, including iterative reconstruction, iterative reconstruction with total variation, and the diffusion model alone in terms of quantitative indices such as the structural similarity index and peak signal-to-noise ratio. We also explored further sparse-projection CT reconstruction using 1/20 projection data with the same trained diffusion model. As the number of iterations increased, the image quality improved comparable to that of 1/10 sparse-projection CT reconstruction. In principle, this method can be widely applied not only to CT but also to other imaging modalities.

\end{abstract}

%\fontsize{11}{14}\selectfont
% keywords can be removed

%\newpage

\section{Introduction}

Computed tomography (CT) is an innovative technology used for imaging the interior of the human body~\cite{Hounsfield1973, Buzug2008}.
In clinical practice, CT is used for radiodiagnosis and  radiotherapy. There are several demands for improving the quality of CT images. In CT imaging, image quality and radiation exposure have a tradeoff relationship. As the radiation dose increases, more information can be obtained and, thus, the image quality can be improved. On the one hand, increased radiation exposure poses significant health risks. On the other hand, as the radiation dose decreases, the image quality deteriorates. Ideally, we aim to obtain high-quality images while reducing radiation exposure. This is particularly important for CT imaging in children, where it is crucial to maintain sufficient image quality for diagnosis while minimizing radiation exposure. CT images are also used in radiotherapy, such as treatment planning and patient positioning. Specifically, the quality of CT images used for patient positioning in image-guided radiotherapy (IGRT) is poor. This image quality issue affects the treatment accuracy of IGRT.

Mathematically, the imaging process of CT is an inverse problem. In general, the reconstruction problems of low-quality CT images, as mentioned above, are highly ill-posed. Several methods have been investigated to overcome this ill-posedness and improve the quality of CT images. These methods can be broadly categorized into two types: data-free methods and modern deep learning--based approaches. In data-free methods, iterative reconstruction (IR) is often used to improve the image quality and reduce the radiation dose~\cite{Geyer2015, Beister2012, Katsura2012, Willemink2013, Ozaki2019}.
An advantage of IR is that it can incorporate prior information about an image in the form of regularization terms, such as total variation (TV) regularization~\cite{Candes2006a, Candes2006, Sidky2006}.
Regularization terms constructed based on prior information constrain image optimization, thereby mitigating ill-posedness.
D.~O.~Baguer et al.~\cite{Baguer2020} explored CT reconstruction using a deep image prior (DIP) \cite{Ulyanov2020}. 
In this method, a deep neural network is integrated into IR, and the network parameters are updated instead of the image. Because the functional form of the neural networks is restricted, DIP serves as a regularization approach.

There are numerous applications of deep learning--based methods for improving the image quality in medical imaging. However, these applications can be viewed from a unified perspective, namely, image-to-image translation~\cite{Kaji2019}.
To enhance the image quality, deep generative models such as generative adversarial networks (GANs)~\cite{Goodfellow2014} are often employed in image-to-image translation~\cite{Isola2017, Zhu2017}.
A typical application is the translation between two image domains from low-quality images to high-quality images~\cite{Yang2018, Yi2018, Kang2019, Harms2019, Bera2021, Vinas2021, Huang2022}.
In this approach, images are reconstructed from the projection data of both low- and high-quality CT images using typical reconstruction methods such as filtered back projection (FBP) or IR. 
Subsequently, deep learning models learn to map low-quality images to high-quality images.
However, it is difficult to prepare paired low- and high-quality images in general clinical practice. In such cases, unpaired image-to-image translation methods such as cycle-GAN~\cite{Zhu2017} are often employed.
When applying unpaired image-to-image translation to medical images, it is crucial to preserve the anatomical structures because generative models such as GANs often alter the structures of important organs. To address this problem, anatomical structure preservation has been incorporated into the cycle-GAN~\cite{Kida2020a, Ozaki2022}.
Although deep learning--based image-to-image translation methods show significant improvements in the image quality, these models cannot restore missing information during CT reconstruction.
A second application of image-to-image translation is the translation of different domains, from raw data (e.g., projection) to images. AUTOMAP~\cite{Zhu2018} was proposed to directly map projections onto images. Because projection data are nonlocal, unlike images, convolutional neural networks do not perform well on such data. To resolve this problem, \cite{Zhu2018} implemented fully connected neural networks in the first layer of the encoder. However, direct translation from raw data to images requires a large amount of training data to learn the geometries of imaging systems and the physical processes of reconstruction, in addition to the relationship between low- and high-quality images, which makes the training process expensive.
The third application is a hybrid method that combines IR and deep learning models~\cite{Wu2017, Adler2017, Adler2018, Bubba2019, Li2020, Xiang2021}.
Because the system matrix used in IR contains information regarding the geometry of the CT system, deep learning models do not need to learn the entire imaging process. Trained deep-learning models contain prior information based on a large amount of training data. Specifically, in \cite{Lunz2018}, the authors investigated IR methods combined with GANs. In this approach, the adversarial loss serves as a regularizer in IR.

Generative models have been rapidly developed in recent years. In particular, diffusion models that serve as foundational models for recently developed image-generative artificial intelligence, such as stable diffusion~\cite{Rombach2022} and DALL-E3~\cite{Betker2023}, can produce remarkably high-quality images with rich diversity. Although diffusion models can generate high-quality images, they introduce structural changes caused by diversity along with randomness in the output images. This aspect of diffusion models --- or, more broadly, generative models --- is considered undesirable in medical applications.
To mitigate these structural changes, several studies combined diffusion models with iterative CT reconstruction, incorporating data consistency loss~\cite{Song2022, Chung2022a, He2023a, Xia2023, Chung2023a, Dey2024, Du2024, Xie2024, Song2024, Montes2024, Jiang2024}, similar to the aforementioned hybrid deep learning--based and IR methods. An advantage of combining unconditional diffusion models with IR is that these models can be applied to reconstruct any CT modality to enhance the image quality using a single trained model, as training is conducted using only high-quality images in an unpaired manner.
However, previous studies invariably introduced a parameter to control the tradeoff between image quality enhancement by an unconditional diffusion prior and structure preservation by data consistency. For instance, ~\cite{Song2022, Chung2022a, He2023a, Chung2023a, Dey2024, Du2024, Montes2024} introduced a parameter to balance the unconditional diffusion prior and data fidelity. \cite{Xia2023, Xie2024, Song2024, Jiang2024} combined iterative CT reconstruction with data consistency loss in each step of the reverse process, where the generated image from the previous time step $x_{t}$ was used as the initial image for iterative CT reconstruction. In this case, the number of iterations in iterative CT reconstruction serves as a parameter that controls the balance between the generated image $x_{t}$ by the diffusion model and the image reconstructed by the iterative method.
Introducing such a parameter may compromise image quality, structure preservation, or both. Thus, the parameter tuning is a considerably delicate process and depends on the modality, leading to the loss of generalizability.

\begin{figure*} % picture
    \centering
    %\begin{tabular}{c}
    \begin{minipage}[t]{1.0\hsize}
    \includegraphics[width=0.65\linewidth]{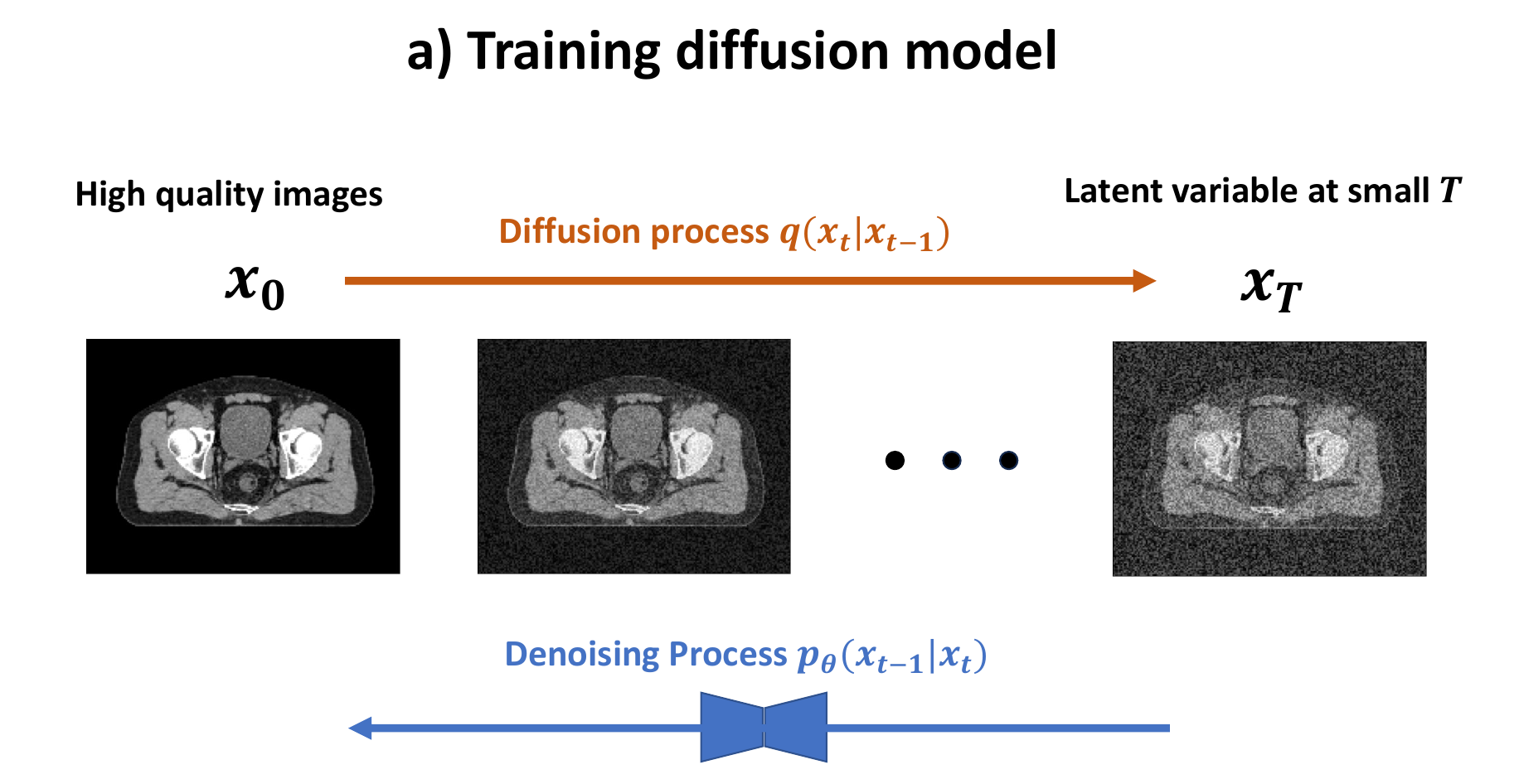}
    \includegraphics[width=0.8\linewidth]{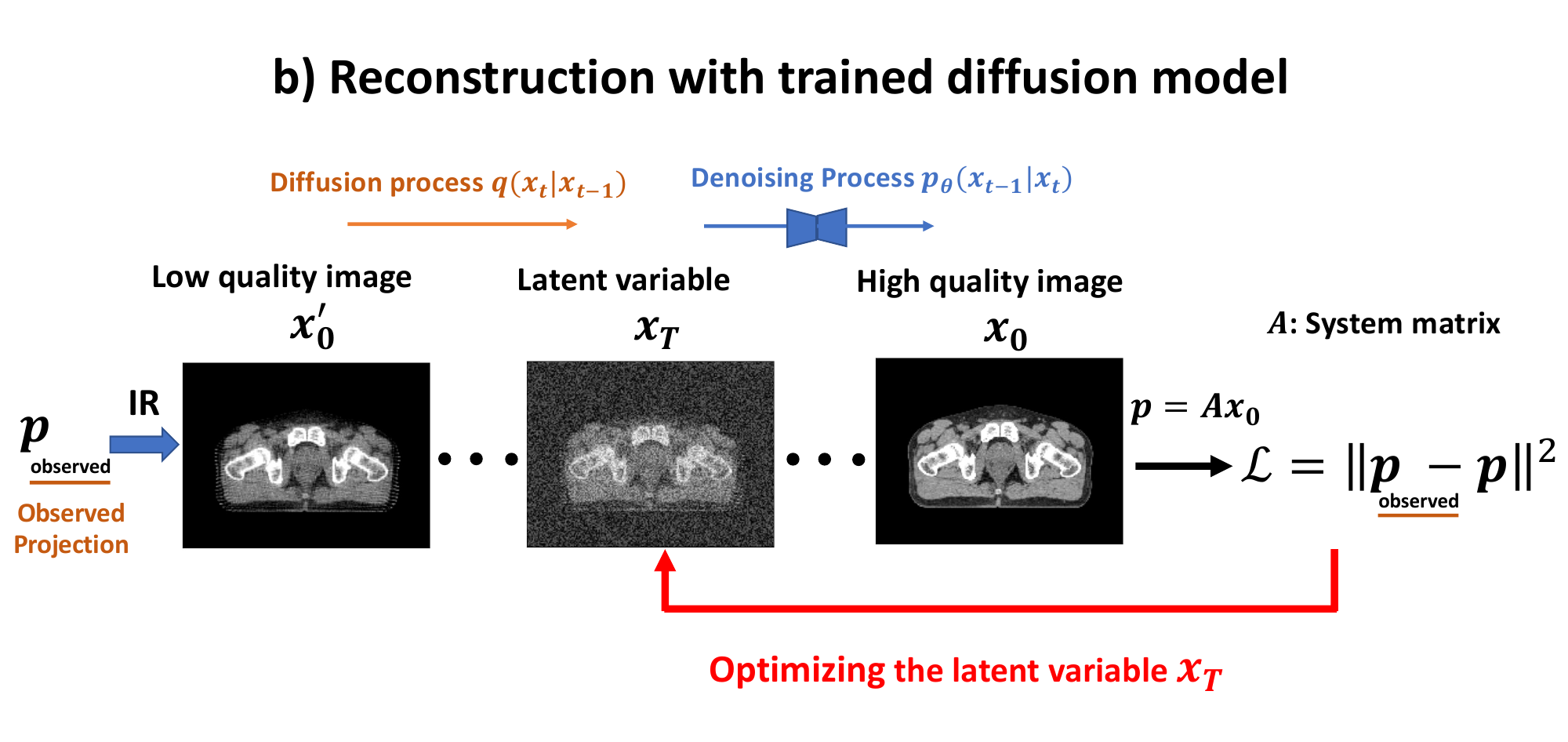}
    \centering
     \includegraphics[width=0.68\linewidth]{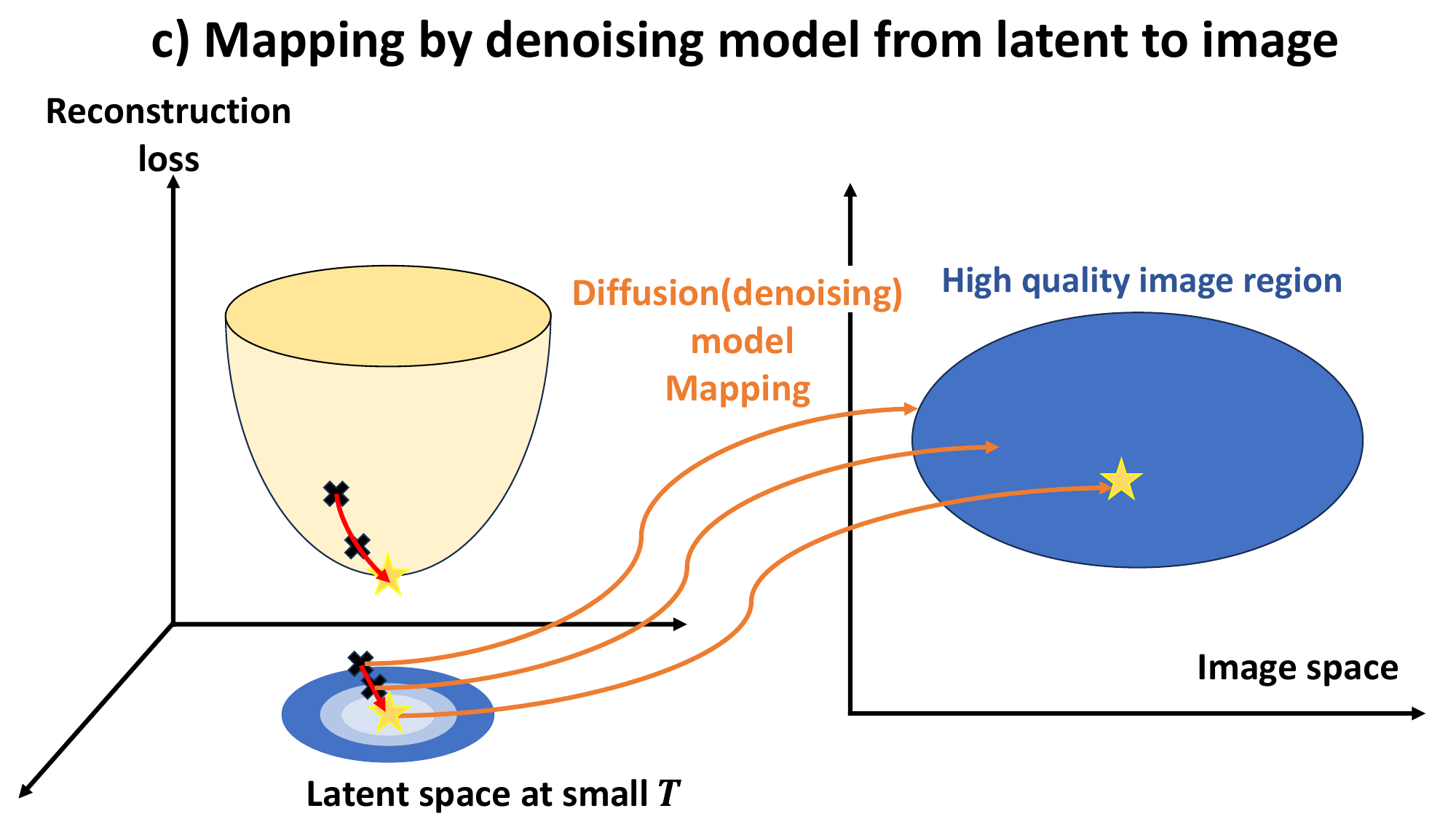}
        \caption{Schema of the proposed method. a) Training of the diffusion model. In the training phase, we stop the diffusion process at a small $T$ such that the latent variable $x_{T}$ can partially contain information on the anatomical structures of a patient. b) Reconstruction using the trained model. At the beginning of the reconstruction phase, we reconstructed an initial image using IR and the observed projection data. Subsequently, we conducted a forward (diffusion) process using the initial image to obtain the latent variable. In the reverse process, noise in the latent variables is subtracted using the trained diffusion model. The final output of the model is a high-quality image. Reconstruction loss is computed using the output and observed projection data, and the loss function is optimized with respect to the latent variable $x_{T}$. c) Mapping from a latent variable to an image using the trained diffusion model. The diffusion model is trained to map the latent variables at a small $T$ to images within the high-quality region. Concurrently, reconstruction loss ensures that the output of the diffusion model preserves the anatomical structure. At the end of the optimization process, a high-quality image is reconstructed while maintaining the anatomical structures of the patient. Notably, the dimensions of the latent space are the same as those of the image space, for example, $256 \times 256$ or $512 \times 512$, although the spaces in the figure are depicted two-dimensionally (2D) for simple visualization.
    \label{Fig: Proposed method.} }
    \end{minipage}
    %\end{tabular}
\end{figure*}

In this study, we proposed a novel method that combines an unconditional diffusion model with IR. In contrast to previous studies, we imposed a consistency loss (reconstruction loss) of projection data on the output of the diffusion model and optimized the reconstruction loss with respect to the latent variable of the diffusion model during the IR process, instead of the image and model parameters. Because the projection data used in CT reconstruction contain information about the anatomical structure of the patient, enforcing consistency loss between the projection data and the output from the diffusion model allows us to preserve the structures in the images without compromising the image quality.
In the proposed method, we do not need to introduce a parameter to control the tradeoff between image quality and reduction of structural changes. We used the denoising diffusion probabilistic model (DDPM) as the diffusion model~\cite{Ho2020, Sohl-Dickstein2015}.
In the training phase of the ordinary DDPM, we gradually increase the noise component of the clean input image while reducing the data component of the image at each increment of the time step $t$ (diffusion process). At the final time step $t=T$, we obtain a perfectly noisy image. In an ordinary DDPM, $T$ is set as $T=1000$. Next, we reverse the process from the noisy image at $t=T$ and recover the original clean image by subtracting the added noise using a denoising neural network (denoising or reverse process).
Therefore, a denoising neural network is trained during the training phase. In the inference (generation) phase, we only conducted the reverse process with the trained denoising model with the initial random noise, creating a new clean image with diversity.
The sources of the randomness of diversity, leading to anatomical structural changes, produced by the DDPM during the inference phase (reverse process) are twofold:
\begin{enumerate}
 \item Initial random noise at $t=T$ 
 \item Random noise added at each timestep 
\end{enumerate} 
To suppress the first source, we stopped the diffusion process at a small $T$ (typically, $T\ll 1000$), avoiding the pure noise image during training, and then started at a small $T$ in the reverse process.
Hence, the latent variable can partially contain information regarding the anatomical structure of the patient. We explored the $T$-dependence of the results. 
The authors of~\cite{Jiang2024} also reported that jumpstart sampling of the reverse process significantly reduces the reconstruction time.
In previous studies, IR with data consistency loss was introduced at each time step of the reverse process to reduce the secondary source of randomness. Instead of introducing IR into the reverse process, the proposed method fixes a set of added noises in the reverse process of the generation (reconstruction) phase. Thus, the denoising model is deterministic and suppresses the second source of random diversity.
The schema of the proposed method is shown in Figure 1.
Figure 1-a) illustrates the training phase of the diffusion model, where we stopped the diffusion process at a small $T$ such that the latent variable $x_{T}$ can partially contain information about the anatomical structure, as mentioned above.
As shown in Figure 1-b), we first reconstructed the initial image using IR with the observed projection data. Subsequently, we conducted a forward diffusion process using the initial image and obtained the initial latent variable. We then optimized the reconstruction loss with respect to the latent variable $x_{T}$. Because the denoising model was trained to map latent variables to high-quality images, we expected that the output of the DDPM would be restricted to the high-quality image region in the image space while preserving the anatomical structures owing to reconstruction loss (see Figure 1-c). Thus, the trained DDPM largely regularizes CT reconstruction in the proposed method.

The remainder of this paper is organized as follows: In Section 2, we describe the construction of the proposed method and provide brief explanations regarding IR and DDPM. In Section 3, we present the experimental results of sparse-projection CT reconstruction by the proposed method, starting with the impact of fixed noise in the reverse process. We also investigated the $T$-dependence of the proposed method. Comparisons between the proposed and existing methods are presented in the final part of Section 3. In Section 4, this approach was tested by exploring further sparse-projection CT reconstruction. Dependence of the proposed method on the noise strength is explored. The limitations of this study are also presented. Finally, in Section 5, we present the conclusions of this study.

\section{Method}

\subsection{Brief description of IR and DDPM}

IR is a well-known approach for CT image reconstruction, characterized by the optimization of data consistency loss, which contains information regarding the anatomical structures and CT imaging geometry. By combining IR with an unconditional DDPM, we can impose structural preservation on the DDPM output. In this study, we combined the DDPM with IR in a different manner from that used in previous studies. First, we briefly explain IR and DDPM to ensure that this section is self-contained.

\subsubsection{IR and total variation regularization}

IR is typically used for ill-posed CT reconstruction tasks.
The basic idea is to solve $Ax=y$, where $y$ represents the observed projection data of each patient and $A$ is the system matrix of the forward CT imaging model containing the geometrical information of the CT system. 
Because this equation is often ill-posed, we formulated the problem of determining $x$ as an optimization problem using regularization terms.
The most commonly used regularization term is TV. The idea is to ensure that the CT values of spatially neighboring pixels do not differ significantly from each other.
TV can reduce noise and streak artifacts; however, it may cause the image to appear overly smooth, resulting in the loss of fine structural details.

Here, we defined the reconstructed image $x=\rm{IR}(y)$ as the minimizer of the following objective function:
\beq
\mathcal{L}
&=& || y - A x ||^2. 
\label{Eq: IR reconstruction loss}
\eeq
With the TV regularization term, the objective function is defined as follows:
\beq
|| y - A x ||^2 + \lambda || \nabla x ||_{1},
\label{Eq: IR reconstruction loss TV}
\eeq
where the parameter $\lambda$ balances the reconstruction loss and TV regularization term.

\subsubsection{DDPM}
Let the initial point $x_0$ be sampled from a manifold consisting of a certain type of data.
The DDPM considers a Markov process called the diffusion or forward process, which adds Gaussian noise:
\beq
q(x_{1:T}| x_0) 
&=& \prod_{t=1}^{T} q(x_t | x_{t-1} ),
\label{Eq: Forward Process}
\eeq
where the conditional probability $q(x_t | x_{t-1} )$ is expressed by the Gaussian distribution as follows:
\beq
q(x_t | x_{t-1} )
&=& \mathcal{N} (x_t; \sqrt{\alpha_t } x_{t-1}, \beta_t I).
\label{Eq: Conditional Probability}
\eeq
Here, $I$ is the identity matrix, and the sequence $0 < \beta_1 < \beta_2 < \cdots < \beta_T \le 1$ is called noise scheduling, which is typically defined as follows:
$\beta_t = t(10^{-2}-10^{-4})/1000$ and $T=1000$,
and $\alpha_t = 1 - \beta_t$.
Using this process, a sequence $x_0,x_1,\ldots,x_T$ that gradually leaves the data manifold and approaches complete noise is obtained.
Owing to the reproductive properties of the Gaussian distribution, a sample at an arbitrary time $x_t ~\sim q(x_t | x_0)$ can be obtained using the following Gaussian distribution:
\beq
q(x_t | x_0)
&=& \mathcal{N}(x_t; \sqrt{ \bar{\alpha}_{t} } x_0, \bar{\beta}_{t} I ),
\label{Eq: ArbitraryGaussian}
\eeq
with
$
\bar{\alpha}_{t} 
= \prod_{s=1}^{t} \alpha_{s}, 
$ and 
$
\bar{\beta}_{t}
= 1 - \bar{\alpha}_{t}. \nonumber
$
With a large $T$, the distributions of $q(x_T | x_0)$ and $q(x_T)$ converge to $\mathcal{N}(x_T; 0, I)$ for an arbitrary $x_0$.

The DDPM also considers the reverse of the Markov process, called the inverse diffusion or reverse process. 
This process converges to a distribution on the data manifold for any initial point sampled from the Gaussian, providing a means to parametrize the data manifold.
Each step of the inverse diffusion process is expressed as a Gaussian distribution as follows:
\beq
p_{\theta} (x_{0:T})
&=& p(x_T) \prod_{t=1}^{T} p_{\theta} (x_{t-1} | x_t ), \\
p_{\theta}(x_{t-1}|x_t)
&=& \mathcal{N}( x_{t-1}; \mu_{\theta}(x_t, t), \beta_t I ), 
\eeq
with 
$
p(x_T)
= \mathcal{N}(x_T; 0, I).
$
The core idea is to approximate the mean value by
\beq
\mu_{\theta}(x_t, t)
&=& \frac{1}{ \sqrt{ \alpha_t } } \left( x_t - \frac{ \beta_t }{ \sqrt{ \bar{\beta}_t } } \epsilon_{\theta} (x_t, t ) \right),
\eeq  
where $\epsilon_{\theta} (x_t, t )$ is a deep neural network model with a learnable parameter $\theta$ that predicts the noise of $x_t$ at $t$.

To learn the noise prediction model $\epsilon_{\theta} (x_t, t )$, we optimized the following objective function:
\beq
\mathcal{L}
&=& \sum_{t=1}^{T} \mathbb{E}_{x_0 \sim P_{\rm{data}}(x_0), \epsilon \sim \mathcal{N}(0,I) } 
\left[ 
\frac{ \beta^2 }{ 2 \sigma_{t}^{2} \alpha_{t} \bar{\beta}_{t} } || \epsilon -\epsilon_{\theta} (x_t, t ) ||^{2}
 \right], \nonumber \\
\label{Eq: Objective DDPM}
\eeq
where $x_{t}$ is sampled using $q(x_t | x_0)$ in Eq. (\ref{Eq: ArbitraryGaussian}) under the condition $x_{0}$ (see Algorithm \ref{Algorithm: Learning}).
Using the reparameterization technique, $x_{t}$ can be expressed by $x_{0} \sim P_{\rm{data}}(x_0)$ and $\epsilon \sim \mathcal{N}(0,I)$ as follows:
\beq
x_{t}
&=& \sqrt{ \bar{ \alpha }_{t}  } x_{0} + \sqrt{ \bar{\beta}_{t} } \epsilon.
\eeq
Using the objective function (\ref{Eq: Objective DDPM}), the model $\epsilon_{\theta} (x_t, t )$ learns to predict the added noise $\epsilon$ in $x_{t}$.
In the inference (image generation) phase, the DDPM uses the learned denoising model $\epsilon_{\theta} (x_t, t )$ and creates a new clean image $x_{0}$ through a reverse process starting from the Gaussian noise $x_{T} \sim \mathcal{N}(x_T; 0, I)$.

\subsection{Iterative CT reconstruction using the deterministic shallow DDPM}

The DDPM creates high-quality images with rich diversity.
Although diversity with randomness is desirable for natural image generation tasks, it is undesirable for medical imaging as it would forge inexistent structures.
To address this problem, we constructed deterministic mapping using the DDPM with a small $T$ (shallow DDPM [SDDPM]) and a fixed set of noises, eliminating the source of diversity in the original DDPM, which is combined with IR.

The key concept is to consider the reverse process as a mapping that parameterizes the data manifold.
To construct the mapping deterministic, 
we fixed a set of noises $\{u_i\mid 1\le i \le T\}$ with $u_1=0$
and defined
\beq
x_{0}
&=& f_{\theta, T, \{u_i\}} (x_{T}).
\eeq
as in lines 5--7 of
Algorithm \ref{Algorithm: Reconstruction}.
The proposed image reconstruction process with the observed projection data $y$ is defined by 
$x = f_{\theta, T,\{u_i\}} (x_{T})$, where
$x_{T}$ is the minimizer of the following objective function:
\beq
\mathcal{L}
&=& || y - A x_0 ||^{2}
= || y - A f_{\theta, T,\{u_i\}}(x_{T}) ||^{2}.
\label{Eq. Objective function for the mapping function}
\eeq
As an effective initial guess, we proposed the following:
\beq
x_{T}^{\prime}
&\sim& q(x_{T}^{\prime} | {\rm{IR}}(y)).
\eeq
In other words, Gaussian noise was added to the image reconstructed using pure IR.
The reconstruction process is presented in Algorithm \ref{Algorithm: Reconstruction}.

The mapping $f_{\theta, T,\{u_i\}}$ can be considered as ``change of variables'' for the optimization.
The mapping is learned to output images with reduced noise. In other words, the output of the mapping is restricted to a high-quality image region in the image space, as mentioned in Introduction and illustrated in Figure 1-c. 
For more clarity, let us consider a simple toy problem.
Let us assume that we want to find a location on a unit circle that minimizes the function $g(x,y)$ defined over the plane. This is a constrained optimization problem; however, 
it can be reduced to an unconstrained problem 
for the univariate function $g(f(\theta))$ through the parametrization $f(\theta)=(\cos{\theta},\sin{\theta})$.
This change of variable is achieved by mapping $f$ from real numbers to the unit circle.
Recently, a similar mapping process has been investigated~\cite{Gutha2024} using consistency models~\cite{Song2023}.

Despite the simplicity of the implementation, the proposed method reconstructs high-quality images while preserving the anatomical structure of the patient, as demonstrated in the next section.

\begin{algorithm}[tb]
\caption{DDPM model Learning algorithm}
\label{Algorithm: Learning}
\begin{algorithmic}[1]
\Require{learning rate $\eta>0$, maximum number of steps $T$, noise scheduling $\{ \beta_{t} \}$ }
   \Repeat
    %\ForAll {$element \gets array$} 
    \State $x_0 \sim P_{\rm{data}}(x_0)$
    \State $t \sim {\rm{Uniform}} ( \{ 1, 2, \cdots, T \} )$
    \State $\epsilon \sim \mathcal{N}(0, I)$
    \State $x_{t} = \sqrt{ \bar{ \alpha }_{t}  } x_{0} + \sqrt{ \bar{\beta}_{t} } \epsilon$
    %\State $\Delta \theta = \nabla_{\theta}  || \epsilon -\epsilon_{\theta} (x_t, t ) ||^{2}$ 
    \State $\theta = \theta - \eta \nabla_{\theta}  || \epsilon -\epsilon_{\theta} (x_t, t ) ||^{2}$
   \Until{ converged }

\end{algorithmic}
\end{algorithm}

% ref
%ref test \ref{Algorithm: Learning}

\begin{algorithm}[tb]
\caption{Reconstruction algorithm}
\label{Algorithm: Reconstruction}
\begin{algorithmic}[1]
\Require{learning rate $\gamma>0$, projection $y$, maximum number of steps $T$, noise scheduling $\{ \beta_{t} \} $}
    \State $z \sim q(x_{T}^{\prime} | {\rm{IR}}(y) )$ 
	\State $u_1 \gets 0$
	\State $u_2,u_3,\ldots,u_t \sim \mathcal{N}(0,I)$ \, \,  [Fixed set of noises]
    % \State $x_0^{\prime} = {\rm{IR}}(y)$
    % \State $x_{\hat{t}}^{\prime} \sim q(x_{\hat{t}}^{\prime} | x_{0}^{\prime} )$
    % \State $x_{\hat{t}} \gets x_{\hat{t}}^{\prime}$
    % \State $z \sim q(x_{T}^{\prime} | {\rm{IR}}(y) )$
    \Repeat
        \State $x_{T} \gets z$
    	\For{$t = T, T-1, \cdots, 1$}
		% \State $u_t \sim \mathcal{N}(0,I)$  \, [seed fixed]
		% \If{$t=1$} 
		% 	\State $u_{t}=0$
		% \EndIf
		\State $x_{t-1} = \frac{1}{\sqrt{\alpha_{t}}} \left( x_{t} - \frac{ \beta_{t} }{ \sqrt{ \bar{\beta}_{t}} } \epsilon_{\theta} (x_{t}, t) \right) + \sqrt{\beta_{t}} u_{t} $
	\EndFor
	%\State $\mathcal{L} = || y - A x_{0} ||^2 $
	\State $z = z - \gamma \nabla_{z} || y - A x_{0} ||^2$
   \Until{ converged }

\end{algorithmic}
\end{algorithm}

\section{Experiments}

\subsection{Experimental setup}

We demonstrated the effectiveness of the proposed method through sparse-projection CT reconstruction of the lower abdominal regions, where we reduced the number of projection data at equal intervals,
and explored how much lost information is restored in the reconstructed image.
We used a dataset of high-quality planning CT images for radiation therapy, consisting of 3615 images (2D slices) from 20 patients with prostate cancer who underwent stereotactic radiotherapy at the University of Tokyo Hospital.
These images were acquired using a helical CT scanner (Aquilion LB, Canon Medical Systems, JP) with a tube voltage of 120 kV and a tube current of 350 mA.
The pixel size of the image was $1.074$ $\times$ $1.074$ mm, with a slice thickness of $1.0$ mm.
Owing to GPU memory limitations, we downsampled all images from the original size of $512 \times 512$ pixels with a pixel spacing of $1.074$ mm to $256 \times 256$ pixels with a pixel spacing of $2.148$ mm.
These images served as the training dataset for the SDDPM. 
We further collected 20 slices from 10 test patients (two slices per patient, with a spacing of more than 40 slices apart) and downsampled them to $256 \times 256$ pixels in the same manner as the training dataset.
These test images were used for the evaluation.
We simulated CT measurements (projection data) with a fan-beam geometry.
The imaging geometry used in this study was as follows: the source-to-center distance was $115.0$ cm and the source-to-detector distance was $177.2$ cm.
The arc detector array had 528 elements with a detector spacing of $1.250$ mm, and 800 samples were acquired during gantry rotation.

Using the 20 test images, full-scan CT reconstructions was performed using IR, and sparse-projection (1/10 projection data) CT reconstruction was performed using IR, IR+TV, and the proposed method.
To quantitatively evaluate the image quality, we employed the structural similarity index (SSIM) and peak signal-to-noise ratio (PSNR) metrics, with the full-scan CT image reconstructed using IR serving as the ground truth.
First, we explored the effects of fixed noise and searched for an optimal time step $T$ using the proposed method.
We then compared the proposed method with other methods including IR, IR+TV, and SDDPM alone.
For IR+TV, we set $\lambda = 2 \times 10^{-4}$ such that the median values of SSIM and PSNR were maximized when reconstructing the 20 test images using IR+TV.

All training cycles were conducted using a single NVIDIA Tesla V100 GPU, whereas reconstructions with the trained models were performed using a single NVIDIA RTX 2080 Ti GPU.
The proposed algorithm was implemented with Python and PyTorch.
For efficiency, the intensity of CT images was clipped to [$-500, 200$] HU and scaled to [$-1, 1$].
Denoising networks were trained from scratch at a learning rate $ \eta = 10^{-4}$ using the Adam optimizer and a batch size of 10.
We used 500 epochs for all training cycles.
During the reconstruction stage, we used the SGD optimizer with $\gamma = 10^{4}$ and set the number of iterations to $3000$.

Regarding noise scheduling for the SDDPM, we used the cosine schedule, which is an improved schedule proposed in ~\cite{Nichol2021}. The cosine schedule is defined in terms of $\bar{\alpha}_{t}$ as follows:
\beq
\bar{\alpha}_{t} = \frac{g(t)}{g(0)}, \ \ \ \ \ \ g(t) = {\rm{cos}} \left( \frac{t/1000 + s}{ 1+s} \cdot \frac{\pi}{2} \right)^{2},
\label{Eq: Cosine Schedule}
\eeq
where $s$ is set to $s=0.008$ in the original study~\cite{Nichol2021}.
Using $\bar{\alpha}_{t}$ from Eq.~(\ref{Eq: Cosine Schedule}), $\beta_{t}$ is defined as follows:
\beq
\beta_{t} 
&=& 1 - \frac{ \bar{\alpha}_{t} }{ \bar{\alpha}_{t-1}}.
\eeq

\subsection{Effect of fixed noises}

\begin{figure*} % picture
    \centering
    %\begin{tabular}{c}
    %\begin{minipage}{1.0\hsize}
    %\fbox{\rule[-.5cm]{4cm}{4cm} \rule[-.5cm]{4cm}{0cm}}
    %\includegraphics[width=1.0]{fig1.png}
    \includegraphics[width=0.6\linewidth]{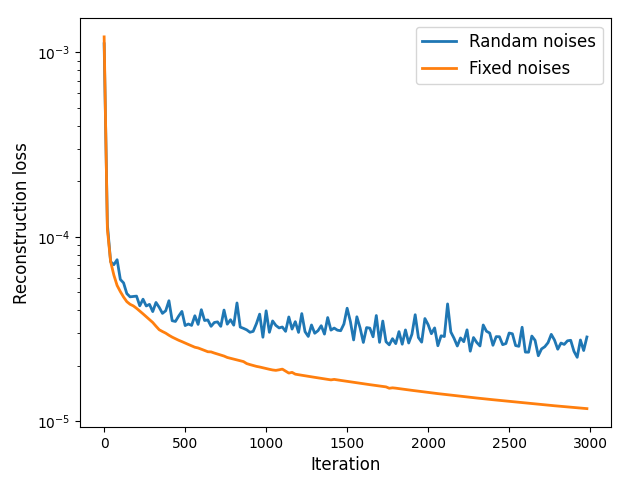}
    \caption{Reconstruction loss as a function of iterations. The losses of the proposed method with fixed and random noises are shown. In this analysis, $T=100$ is used for the SDDPM.
    \label{Fig: ReconstructionLoss_timestep100} }
    %\end{minipage}
    %\end{tabular}
\end{figure*}

\begin{figure*} % picture
    \centering
    %\begin{tabular}{c}
    %\begin{minipage}{1.0\hsize}
    %\fbox{\rule[-.5cm]{4cm}{4cm} \rule[-.5cm]{4cm}{0cm}}
    %\includegraphics[width=1.0]{fig1.png}
    \includegraphics[width=1.0\linewidth]{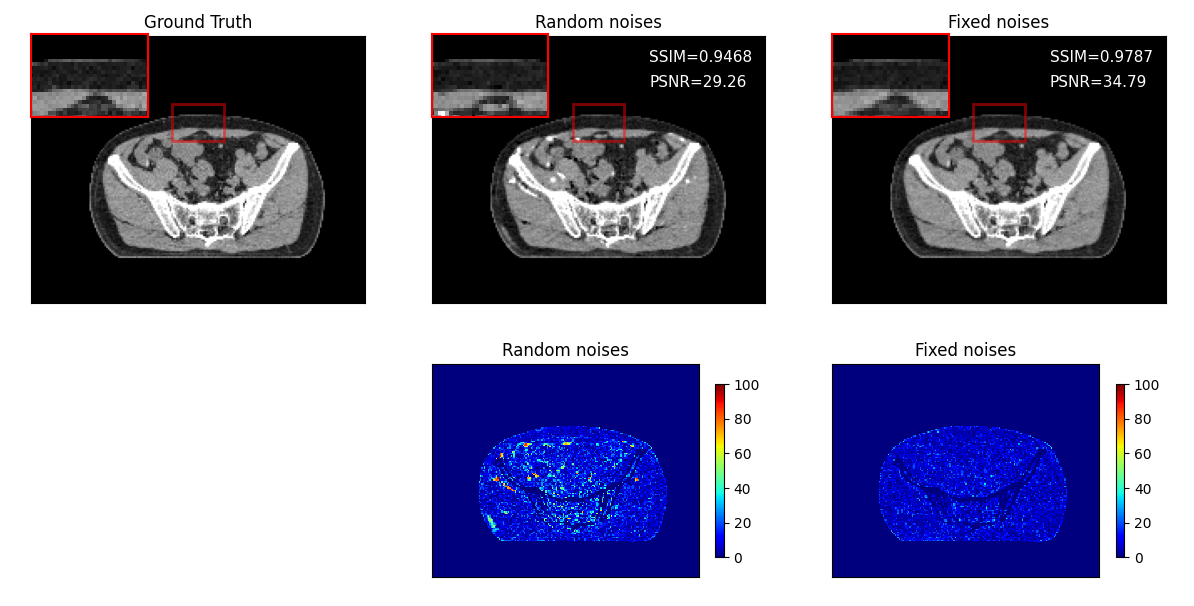}
    \caption{Visual comparison of images reconstructed using the proposed method with fixed and random noises. The upper row shows the images reconstructed using the proposed method with fixed and random noises and the ground truth image. 
    The ground truth image is reconstructed by IR using full scan projection, whereas other images are reconstructed by the proposed method using a 1/10 sparse projection.
    A display window of [-150, 200] HU is used for each image.
    The values of SSIM and PSNR are indicated.
    The lower row shows the absolute value of the difference between each image and the ground truth image. A display window of [0, 100] HU is used for the heat maps.   
    \label{Fig: VisualCompTimeStep100} }
    %\end{minipage}
    %\end{tabular}
\end{figure*}

To assess the impact of fixed noises on the proposed method, we compared the results obtained using fixed noises with those obtained using random noises.
We set $T=100$, which may be sufficiently small compared with $T=1000$ ($T$-dependence of the results will be explored in the next subsection).
Figure~\ref{Fig: ReconstructionLoss_timestep100} shows reconstruction losses with fixed and random noises.
The loss with fixed noises rapidly decreased compared with that with random noises.
This illustrates an advantage of using fixed noises in the proposed method.
Another advantage is evident in the reconstructed images.
Figure~\ref{Fig: VisualCompTimeStep100} provides a visual comparison of fixed and random noise cases with the ground truth image.
All images were reconstructed using 1/10 sparse-projection data, except for the ground truth image.
In the fixed noise case, a clean image was obtained using the proposed method.
By contrast, in the image with random noise, higher signal artifacts appeared, which did not exist in the ground truth image.
This is attributed to the diversity generated by randomness. 
We also observed that the anatomical structures slightly changed in the image with random noise.
Consequently, the index values for the random noise case were lower than those for the fixed noise case. Higher signal artifacts and structural changes are also evident in the heat maps shown in Fig.~\ref{Fig: VisualCompTimeStep100}, highlighting the differences between each image and the ground truth image. Based on these results, we used a fixed set of noises in the proposed method for subsequent analyses.

\subsection{Variations in the maximum number of time steps $T$}

\begin{figure*} % picture
    \centering
    %\begin{tabular}{c}
    %\begin{minipage}{1.0\hsize}
    %\fbox{\rule[-.5cm]{4cm}{4cm} \rule[-.5cm]{4cm}{0cm}}
    %\includegraphics[width=1.0]{fig1.png}
    \includegraphics[width=0.65\linewidth]{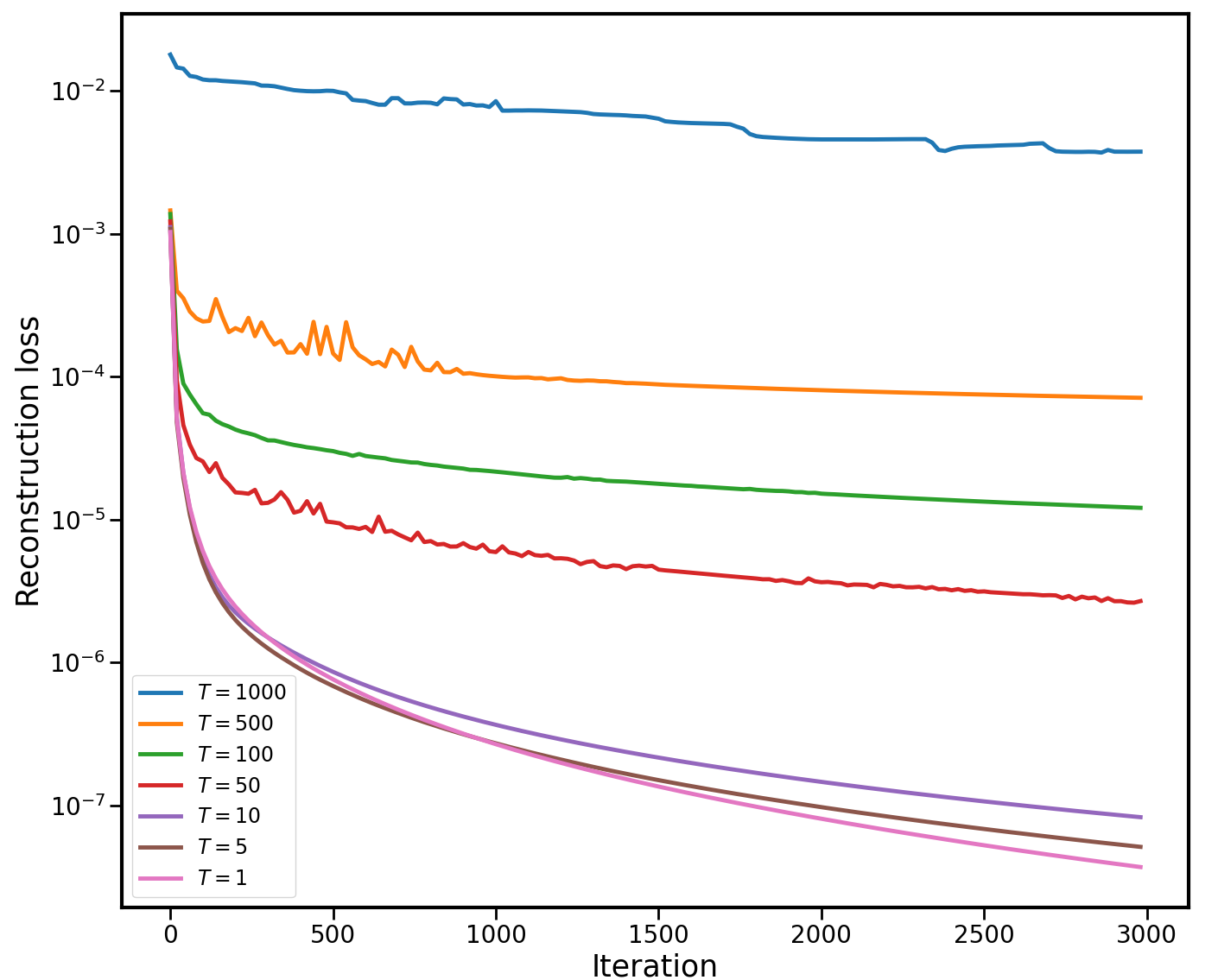}
    \caption{Reconstruction loss as a function of iterations. Losses from the proposed method with 
    $T=1, 5, 10, 50, 100, 500$, and $1000$ are shown.
    \label{Fig: ReconstructionLoss} }
    %\end{minipage}
    %\end{tabular}
\end{figure*}

\begin{figure*} % picture
    \centering
    %\begin{tabular}{c}
    %\begin{minipage}{1.0\hsize}
    %\fbox{\rule[-.5cm]{4cm}{4cm} \rule[-.5cm]{4cm}{0cm}}
    %\includegraphics[width=1.0]{fig1.png}
    \includegraphics[width=1.1\linewidth]{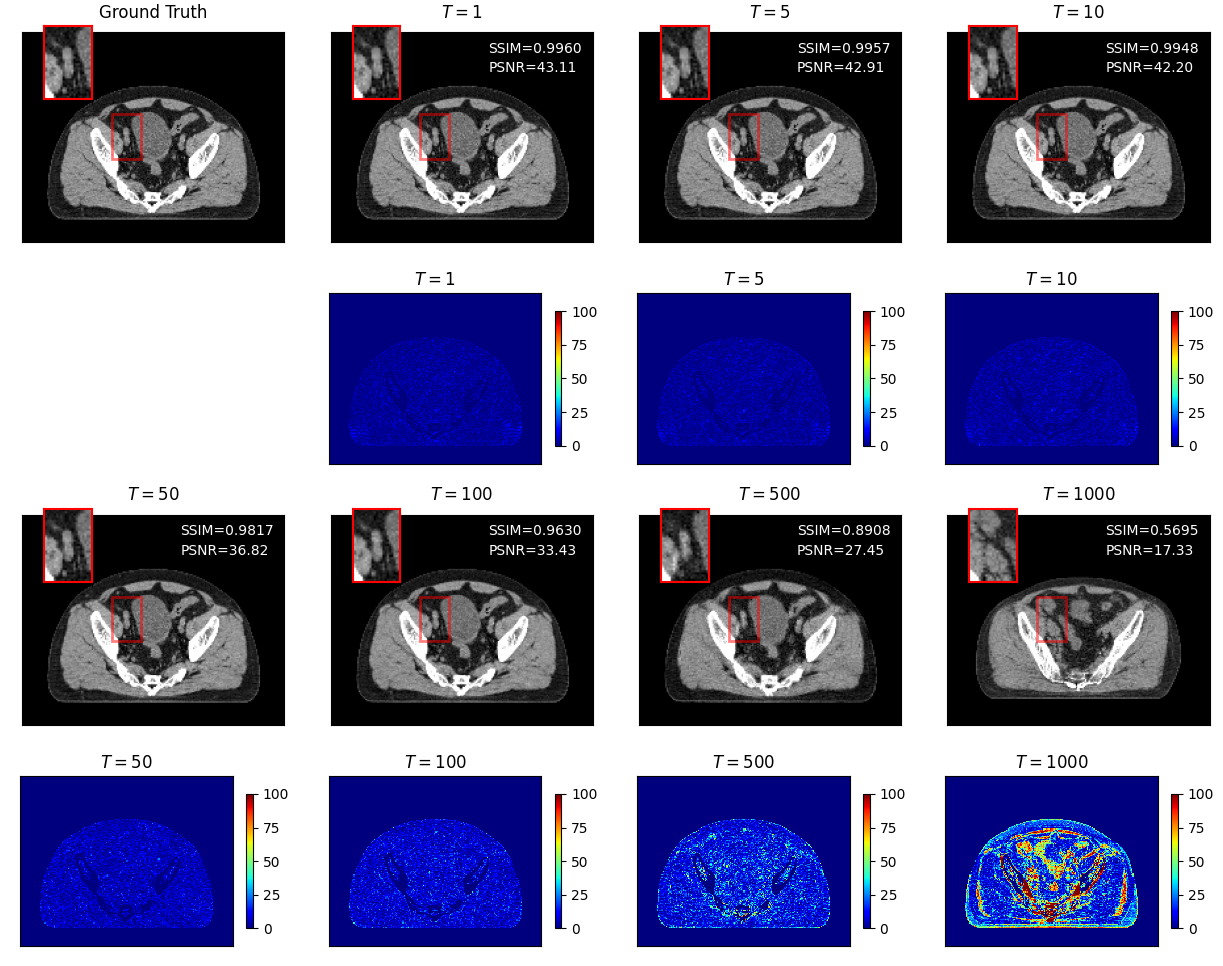}
    \caption{Visual comparison of the images reconstructed using the proposed method with $T=1, 5, 10, 50, 100, 500$, and $1000$. The first and third rows show the ground truth image reconstructed by IR using full scan projection and the images reconstructed by the proposed method using a 1/10 sparse projection.
    A display window of [-150, 200] HU is used for each image.
    The values of SSIM and PSNR are indicated.
    The second and fourth rows show the absolute values of the differences between each image and the ground truth image. A display window of [0, 100] HU is used for the heat maps. 
    \label{Fig: VisCompSSIM_PSNR} }
    %\end{minipage}
    %\end{tabular}
\end{figure*}

\begin{figure*} % picture
    \centering
    %\begin{tabular}{c}
    \begin{minipage}{1.0\hsize}
    %\fbox{\rule[-.5cm]{4cm}{4cm} \rule[-.5cm]{4cm}{0cm}}
    %\includegraphics[width=1.0]{fig1.png}
    %\includegraphics[width=0.5\linewidth]{./figs/ssim_with_several_ts.png}
    \includegraphics[width=0.50\linewidth]{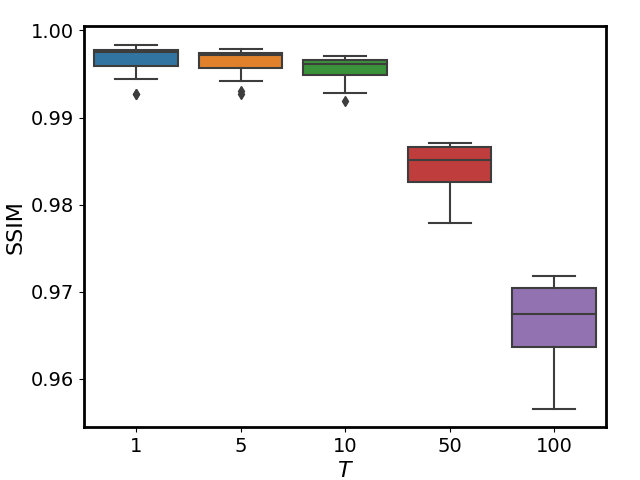}
    \includegraphics[width=0.50\linewidth]{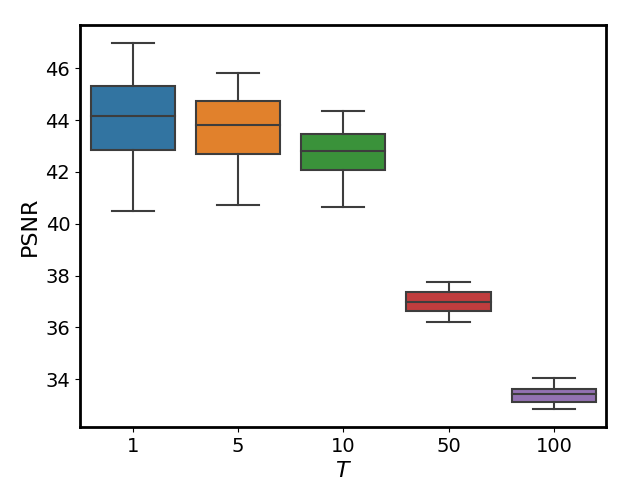}
    \caption{SSIM (left) and PSNR (right) evaluated using the proposed method with $T = 1, 5, 10, 50$, and $100$. Multiple slices (N=20) from test patients are used for the evaluation.
    \label{Fig: Multi-slice_SSIM_PSNR} }
    \end{minipage}
    %\end{tabular}
\end{figure*}

Here, we explored the $T$-dependence of the results of the proposed methods.
The maximum number of time steps $T$ indicates how shallow the SDDPM is.
Notably, the reconstruction time decreased as $T$ decreased
because the time per iteration of the reconstruction decreased with a smaller $T$ in the SDDPM.
Figure \ref{Fig: ReconstructionLoss} shows the $T$-dependence of reconstruction losses.
The reconstruction loss rapidly decreased as $T$ decreased.
In particular, beyond $1000$ iterations, the reconstruction loss for $T=1$ was consistently the lowest.
Figure \ref{Fig: VisCompSSIM_PSNR} presents a visual comparison among several values of $T$ with the ground truth image.
The image with $T=1000$\footnote{in this case, the model is the DDPM} showed significant changes in the anatomical structures compared with the original image (ground truth).
The image with $T=500$ showed slight structural changes; thus, the index values were considerably low compared with those of other images with $T \ge 100$.
The structural changes are also evident in the heat maps shown in Fig.~\ref{Fig: VisCompSSIM_PSNR}.
In the single-slice analysis, the SSIM and PSNR values for the image with $T=1$ were the highest.  
The results of the multi-slice analysis ($N=20$) are shown in Figure \ref{Fig: Multi-slice_SSIM_PSNR}.
As $T$ increased, both the SSIM and PSNR values decreased.
The median values of SSIM and PSNR were the highest when $T=1$.
Therefore, based on these results, we conclude that $T=1$ is the optimal value for the maximum number of time steps in the proposed method.
These results are reasonable because the latent variable with $T=1$ can contain the maximum information regarding the anatomical structure of the patient.

\subsection{Comparison of the proposed method with existing methods}

\begin{figure*} % picture
    \centering
    %\begin{tabular}{c}
    %\begin{minipage}{1.0\hsize}
    %\fbox{\rule[-.5cm]{4cm}{4cm} \rule[-.5cm]{4cm}{0cm}}
    %\includegraphics[width=1.0]{fig1.png}
    %\includegraphics[width=1.1 \linewidth]{./figs/Visual_comp_IR_TV_Diff.png}
    \includegraphics[width=1.0 \linewidth]{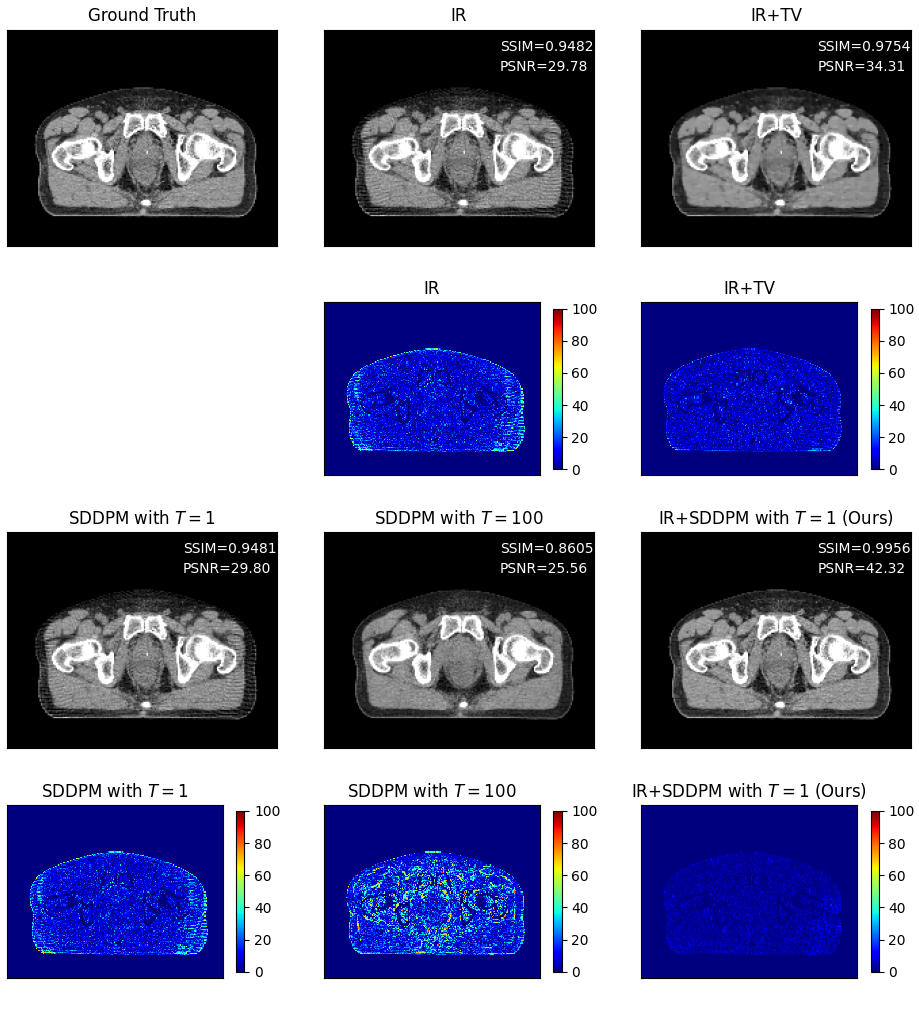}
    \caption{Visual comparison of the images obtained using the proposed and existing methods. The first row shows the ground truth image and the images reconstructed using IR and IR+TV.
    The third row shows the images generated using the SDDPM alone with $T=1$ and $100$, and the image reconstructed using the proposed method with $T=1$.
    The ground truth image is reconstructed by IR using full scan projection, whereas other images are reconstructed using a 1/10 sparse projection.
    The input image of the SDDPM alone with $T=1$ and $100$ is reconstructed by IR using a 1/10 sparse projection.
    A display window of [-150, 200] HU is used for each image.
    The second and fourth rows show the absolute values of the differences between each image and the ground truth image. A display window of [0, 100] HU is used for the heat maps.   
    \label{Fig: VisualCompOthers} }
    %\end{minipage}
    %\end{tabular}
\end{figure*}

\begin{figure*} % picture
    \centering
    %\begin{tabular}{c}
    \begin{minipage}{1.0\hsize}
    %\fbox{\rule[-.5cm]{4cm}{4cm} \rule[-.5cm]{4cm}{0cm}}
    %\includegraphics[width=1.0]{fig1.png}
    %\includegraphics[width=0.5\linewidth]{./figs/ssim_recons.png}
    \includegraphics[width=0.53\linewidth]{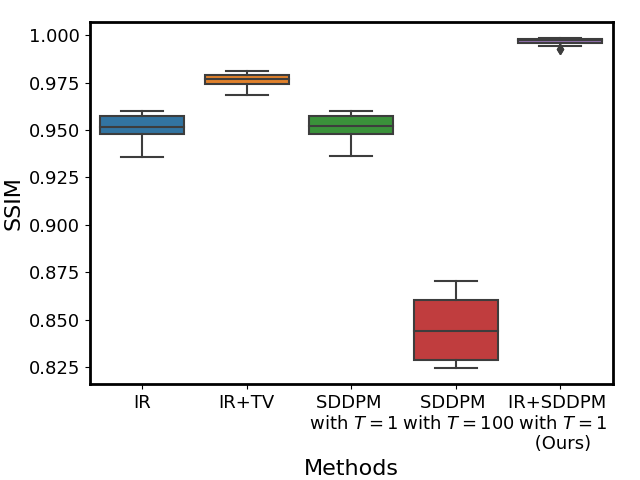}
    \includegraphics[width=0.53\linewidth]{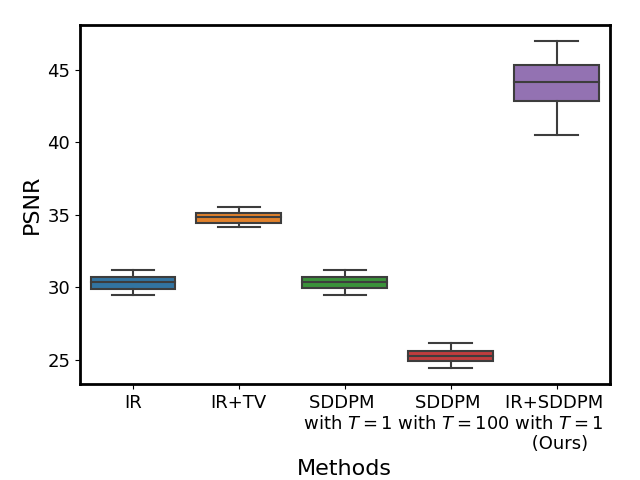}
    \caption{SSIM (left) and PSNR (right) evaluated using the proposed and existing methods. Multiple slices (N=20) from test patients are used for the evaluation.
    \label{Fig: SSIM_PANR_several_reconstrucitons} }
    \end{minipage}
    %\end{tabular}
\end{figure*}

We compared the performance of the proposed method with that of existing methods, including IR, IR+TV, and SDDPM alone.
Figure \ref{Fig: VisualCompOthers} shows a visual comparison between the proposed and existing methods, along with the index values of SSIM and PSNR.
The image reconstructed using IR still exhibited streak artifacts owing to sparse-projection CT reconstruction using 1/10 projection data.
These streak artifacts were eliminated in the image reconstructed using IR+TV; however, some blurring was observed in this case.
The image produced by the SDDPM alone with $T=1$ also exhibited streak artifacts, and the index values were almost the same as those of IR.
For $T=100$, the SDDPM created a clean image at first glance; however, upon closer inspection, slight changes in the anatomical structures compared with the ground truth image were observed. 
Thus, the SSIM and PSNR index values for the SDDPM with $T=100$ were significantly lower than those of the other methods.
The structural changes and streak artifacts are visible in the heat maps shown in Fig.~\ref{Fig: VisualCompOthers}.
In contrast to the existing methods, the proposed method (IR+SDDPM with $T=1$) generated a high-quality image while preserving the anatomical structures of the patient.
Multi-slice analysis ($N=20$) results are shown in Figure \ref{Fig: SSIM_PANR_several_reconstrucitons}.
This figure clearly shows that the proposed method outperforms the other existing methods.

\section{Discussion}

\subsection{Further sparse-projection CT reconstruction using the proposed method}

\begin{figure*} % picture
    \centering
    %\begin{tabular}{c}
    %\begin{minipage}{1.0\hsize}
    %\fbox{\rule[-.5cm]{4cm}{4cm} \rule[-.5cm]{4cm}{0cm}}
    %\includegraphics[width=1.0]{fig1.png}
    %\includegraphics[width=1.1 \linewidth]{./figs/Visual_comp_IR_TV_Diff.png}
    \includegraphics[width=1.05 \linewidth]{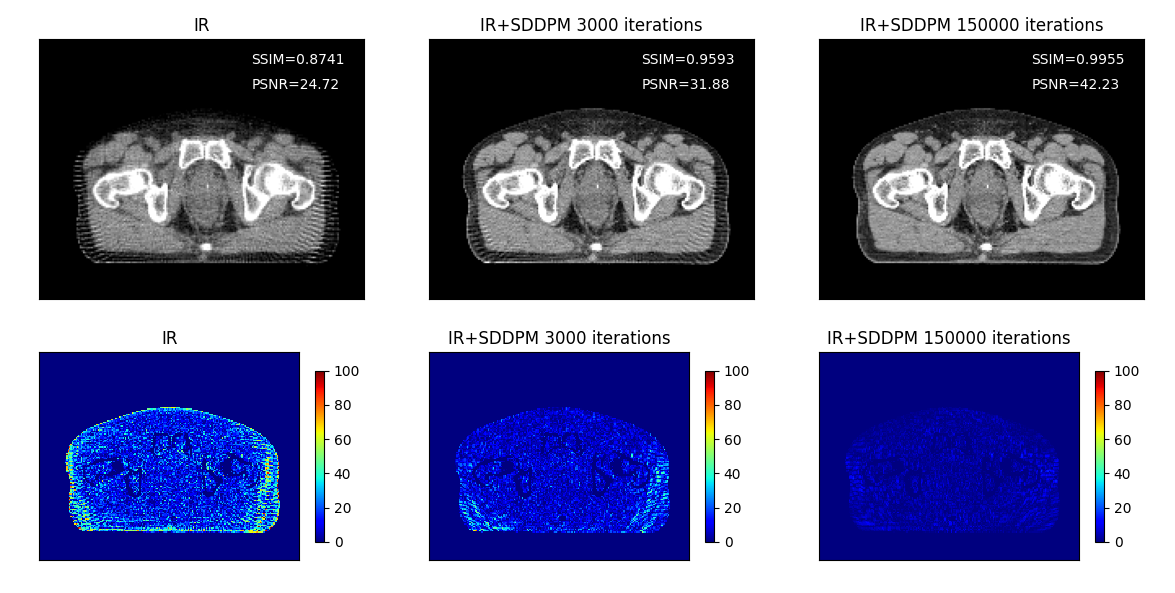}
    \caption{Visual comparison of the images reconstructed using IR and the proposed method with $T=1$ using a 1/20 sparse projection. The first row shows the images reconstructed using IR and the proposed method with 3000 and 150,000 iterations.
    The reconstructed slice is the same as that used in Fig. \ref{Fig: VisualCompOthers}.
    A display window of [-150, 200] HU is used for each image.
    The second row shows the absolute value of the difference between each image and the ground truth image shown in Fig. \ref{Fig: VisualCompOthers}. A display window of [0, 100] HU is used for the heat maps.  
    \label{Fig: Generalization} }
    %\end{minipage}
    %\end{tabular}
\end{figure*}

The learning process of the SDDPM requires only high-quality images and does not depend on projection data.
Therefore, a single trained SDDPM can be used in our scheme to reconstruct not only complete projection data but also sparse or degenerate data.
To test this, we explored the sparse-projection CT reconstruction of 1/20 projection data using the same SDDPM model that was used to reconstruct the 1/10 projection data described in the previous section.
Figure \ref{Fig: Generalization} shows the images reconstructed from the 1/20 projection of the same slice shown in Figure \ref{Fig: VisualCompOthers}.
After 3000 iterations, as in the previous analysis of the 1/10 projection data, the artifacts remained prominent.
However, if the number of iterations was increased to 150,000, the artifacts were significantly reduced and the index values were comparable with those of the proposed method using 1/10 projection data.

\subsection{Dependence on noise scheduling}

\begin{figure*} % picture
    \centering
    %\begin{tabular}{c}
    \begin{minipage}{1.0\hsize}
    %\fbox{\rule[-.5cm]{4cm}{4cm} \rule[-.5cm]{4cm}{0cm}}
    %\includegraphics[width=1.0]{fig1.png}
    %\includegraphics[width=0.5\linewidth]{./figs/ssim_with_several_ts.png}
    \includegraphics[width=0.50\linewidth]{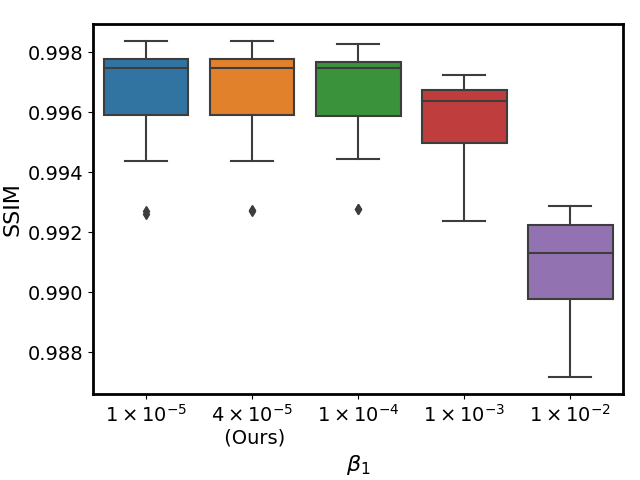}
    \includegraphics[width=0.50\linewidth]{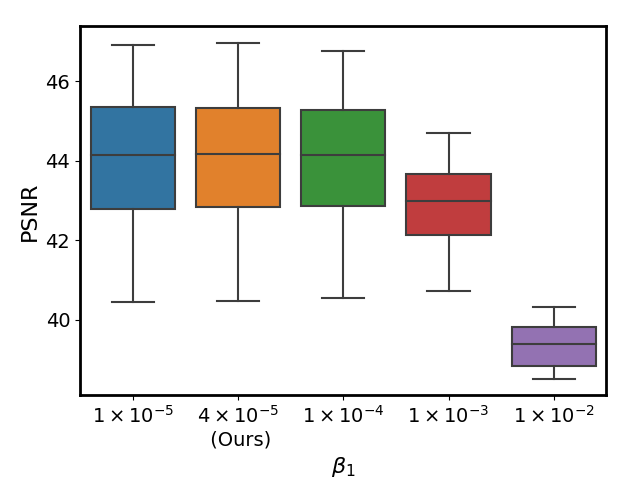}
    \caption{$\beta_{1}$-dependence of the SSIM (left) and PSNR (right) evaluated using the proposed method with $T = 1$. Multiple slices (N=20) from test patients are used for the evaluation. "Ours" refers to the $\beta_{1}$ value in the case of the cosine schedule.
    \label{Fig: beta-dep_SSIM_PSNR} }
    \end{minipage}
    %\end{tabular}
\end{figure*}

We showed that $T=1$ yielded optimal results in terms of SSIM and PSNR. We now investigate how these results vary with changes in the noise strength. For $T=1$, the noise strength is governed by the parameter $\beta_{1}$ in noise scheduling.
The cosine schedule in Eq.~(\ref{Eq: Cosine Schedule}) uses the default value of $\beta_{1} \simeq 4 \times 10^{-5}$. By modifying $\beta_{1}$, we assessed its effect on the SSIM and PSNR.
Figure \ref{Fig: beta-dep_SSIM_PSNR} illustrates a multi-slice analysis ($N=20$) of SSIM and PSNR as functions of $\beta_{1}$. For $\beta_{1} > 1 \times 10^{-4}$, both the SSIM and PSNR exhibited a negative correlation with increasing $\beta_{1}$. However, when $\beta_{1} \le 1 \times 10^{-4}$, the SSIM and PSNR values plateaued, indicating a stable performance.
Our findings suggest that maintaining a sufficiently low noise strength ensures consistent results in the proposed method. The stability in the low-noise regime underscores the robustness of our approach.

\subsection{Limitations}

In this study, we demonstrated the performance of the proposed method by reconstructing images with a resolution of $256 \times 256$ pixels.
However, medical images with a resolution of $512 \times 512$ pixels are typically used in clinical practice.
The limitation of the resolution is attributed to GPU memory.
In addition, the reconstruction speed of methods combining the DDPM with IR is significantly slower than that of the conventional IR.
Our concept of using a shallow DDPM with $T=1$ contributes to both the image quality and computational cost.
However, it still takes a few minutes to reconstruct a slice and approximately 10 h for a volume with a few hundred slices.
Hence, it would not be feasible to use the proposed method for CT reconstruction for patient positioning in radiotherapy because radiation must be delivered immediately after the patient is positioned.
One potential approach to accelerate the reconstruction process by the proposed method is the use of parallelized computation.
Another possible approach is to accelerate the image generation process of the SDDPM.
For example, the authors in Ref.~\cite{Xia2023}  utilized the Nesterov momentum acceleration method to accelerate the DDPM iteration.
This method reduces the computational cost by $20$\% of that of the original method.

\section{Conclusion}

In this study, we proposed a novel CT reconstruction method by combining the DDPM with IR.
Using this method, we optimized reconstruction loss with respect to the latent variable of the trained DDPM.
Unlike previous studies, we did not introduce a parameter to control the tradeoff between image quality and structure preservation.
Furthermore, we identified and suppressed the sources of the randomness of diversity produced by the DDPM by shallowing it (SDDPM) and fixing a set of noises in the reverse process.
Consequently, the proposed method enhanced the image quality while preserving the anatomical structure of the patient. 
We compared the performance of the proposed method with that of other existing methods, such as IR, IR+TV, and DDPM, via 1/10 sparse-projection CT reconstruction. The proposed method outperformed the existing methods in terms of SSIM and PSNR.
We further conducted 1/20 sparse-projection CT reconstruction using the same trained SDDPM.
We found that increasing the number of iterations improves the image quality, which is comparable to that of 1/10 sparse-projection CT reconstruction.
This result opens up the possibility of the general usage of the proposed method with a single trained model for enhancing various CT images, such as those obtained using low-dose CT for pediatric imaging, cone-beam CT, and megavoltage CT used in radiotherapy.
In recent years, magnetic resonance imaging (MRI) reconstruction combined with diffusion models has been actively investigated \cite{Song2022, Chung2022, Chung2023}. In principle, we expect that the proposed method can also be applied to MRI and other medical imaging modalities, including positron emission tomography and single-photon emission CT.

\section*{Acknowledgments}

This research in part used computational resources of Cygnus provided by Multidisciplinary Cooperative Research Program in Center for Computational Sciences, University of Tsukuba.
SK was partially supported by JST Moonshot R\&D Grant Number JPMJMS2021.
TI was partially supported by the JSPS Grant-in-Aid for Scientific Research (C), 21K12121.

\bibliographystyle{ieeetr}

%\bibliography{reference}
\bibliography{IterativeCTReconstructionDiffusionModel}

\end{document}